\documentclass{article}

\usepackage{hyperref}

\usepackage[accepted]{icml2024}

\usepackage{microtype}
\usepackage{graphicx}
\usepackage{subfigure}
\usepackage{booktabs} 
\usepackage{multirow}
\usepackage{bold-extra}

\usepackage{hyperref}

\usepackage{amsmath}
\usepackage{amssymb}
\usepackage{mathtools}
\usepackage{amsthm}
\usepackage{siunitx}
\usepackage{gensymb} 

\usepackage[capitalize,noabbrev]{cleveref}

\theoremstyle{plain}

\theoremstyle{definition}

\theoremstyle{remark}

\newcommand{\tsub}[1]{\textsubscript{#1}}

\DeclarePairedDelimiter\floor{\lfloor}{\rfloor}

\icmltitlerunning{Forecasting Smog Clouds With Deep Learning}

\begin{document}

\twocolumn[
\icmltitle{Forecasting Smog Clouds With Deep Learning: A Proof-Of-Concept}

\icmlsetsymbol{equal}{*}

\begin{icmlauthorlist}
\icmlauthor{Valentijn Oldenburg}{neth}
\icmlauthor{Juan Cardenas-Cartagena}{neth,nor}
\icmlauthor{Matias Valdenegro-Toro}{neth}
\end{icmlauthorlist}

\icmlaffiliation{neth}{Bernoulli Institute, University of Groningen, Groningen, The Netherlands}
\icmlaffiliation{nor}{Wisenet Center, University of Agder, Grimstad, Norway}

\icmlcorrespondingauthor{Valentijn Oldenburg}{v.w.oldenburg@student.rug.nl}
\icmlcorrespondingauthor{Juan Cardenas-Cartagena}{j.d.cardenas.cartagena@rug.nl}

\icmlkeywords{Time series, Deep Learning, Pollution, Meteorology}

\vskip 0.3in
]

\printAffiliationsAndNotice{}

\begin{abstract}
In this proof-of-concept study, we conduct multivariate timeseries forecasting for the concentrations of nitrogen dioxide (NO\tsub{2}), ozone (O\tsub{3}), and (fine) particulate matter (PM\tsub{10} \& PM\tsub{2.5}) with meteorological covariates between two locations using various deep learning models, with a focus on long short-term memory (LSTM) and gated recurrent unit (GRU) architectures. In particular, we propose an integrated, hierarchical model architecture inspired by air pollution dynamics and atmospheric science that employs multi-task learning and is benchmarked by unidirectional and fully-connected models. Results demonstrate that, above all, the hierarchical GRU proves itself as a competitive and efficient method for forecasting the concentration of smog-related pollutants.
\end{abstract}

\section{Introduction}
\label{sec:introduction}

Air pollution and smog carry correlations to numerous pervasive health effects. Given
the risks, foreseeing toxic pollutant levels poses a vital challenge that, upon resolution, enacts a framework for life-saving decisions. Data-driven deep learning (DL) methods offer a novel approach to air quality prediction, yet their potential for modelling a combined set of smog-related pollutants with recurrent neural nets (RNNs) remains unexplored.

The phenomenon of air pollution is characterised by the presence of hazardous atmospheric chemicals. Although a number of physical activities (volcanoes, fire, etc.) may release different pollutants, anthropogenic activities are the head cause of environmental air pollution \citep{kampa2008human}. 

Adverse air pollution effects can range from skin irritation and difficulty in breathing to an increased risk of cardiac and respiratory illnesses, cancer, and mortality overall \citep{brunekreef2002air, kampa2008human, wong2010part, orellano2020short}. A recent addition is its direct link to COVID-19 morbidity and severity \citep{zorn2024effects}. As stated in \citet{lim2012comparative}, air pollution ranks high in the general disease burden attributable to environmental factors, with 3.1 million deaths in 2012 and 3.1\% of disability-adjusted life years worldwide.

Fundamentally, the air pollution problem and the extent to which it spreads is evident. Indeed, the air and its contaminants are everywhere and will remain inevitably inherent to human-nature interaction in the future. Positive notions present themselves, nonetheless: (1) humans, being the primary polluters, also possess the opportunity to act as the ``primary purifiers"; and (2) comprehensive fundamental problem knowledge offers positive prospects for further advancing research \cite{vallero2014fundamentals}. 


This problem motivates the development and application of data-driven forecasting models based on multiple neural network architectures, using contaminant and meteorological data to simulate and predict air pollution and smog.
In particular, we consider the modelling of nitrogen dioxide (NO\tsub{2}), ozone (O\tsub{3}), and (fine) particulate matter (PM\tsub{10} \& PM\tsub{2.5}) with various meteorological covariates as a first proof-of-concept (PoC).
By employing these weather-predictive methods, this study aims to contribute incrementally to understanding air pollution dynamics and enhance environmental conditions for improved public health.

\subsection{State-of-the-art}\label{sec:state_of_the_art}

Traditional weather systems have evolved into sophisticated models that approximate real-world weather dynamics with increasing precision \citep{alley2019advances}.
The systems apply numerical weather prediction (NWP), a now ubiquitous, though computationally costly, numerical method grounded on physical first-principles \citep{bauer2015quiet}.
While applying purely natural laws as boundary conditions for predictions is theoretically possible, it presents challenges in practice: the weather system is everywhere and contains numerous complex processes that make it computationally infeasible to provide these predictions with more than a highly simplified, parameterised value. Moreover, the non-linear dynamics, exemplified by the chaotic behaviour of turbulent flow, make predictions at high resolution---spatially, temporally, and/or across variables---a lasting challenge.

The emergence of data-driven methods presents a novel approach to abstracting physical processes embedded in the weather system. Machine learning (ML) models are adept at recognising complex patterns within large datasets with unparalleled efficiency---patterns that may represent relationships and correlations between atmospheric variables and influences not yet understood by traditional physics. 

A recently undertaken application of large-scale DL weather forecasting is FourCastNet by \citet{pathak2022fourcastnet}. FourCastNet generates global forecasts orders of magnitude faster than traditional NWP with comparable or better accuracy. It herewith demonstrates the potential of data-driven methods to make significant progress in weather forecasting without explicitly considering the underlying (known) physical processes and equations. Implications are reducing costs of the traditional NWP and, more importantly, reducing the opportunity cost of inaccurate forecasts. Its scope does not, however, encompass predicting components directly related to air pollution or smog.

More closely related state-of-the-art studies, see \citet{masood2021review} for a review, that distinctly forecast air pollution are \citet{freeman2018forecasting} and \citet{tao2019air}. The former performs a forecast of surface O\tsub{3} levels using a recurrent neural network (RNN) with long short-term memory (LSTM); its approach takes as input hourly-sampled meteorological data and O\tsub{3} itself, outputting a multivariate 72-hour horizon forecast. The latter \citep{tao2019air} highlights a composition of 1D convnets and the bidirectional gated recurrent unit (GRU) for a multivariate short-term prediction of PM\tsub{2.5}. Both studies are consistent and relevant to the purpose of this PoC in that they use RNNs, take meteorological covariates as inputs, and consequently predict air pollution. Nonetheless, as much as O\tsub{3} and PM\tsub{2.5} are influential elements, a more complete air pollution and smog prediction requires consideration of a broader and combined set of air contaminants.

\subsection{Contributions}

Acknowledging the recent developments \citep{masood2021review} and state-of-the-art, the LSTM and GRU establish themselves as the appropriate choice for modelling the sequential series of components in ambient (polluted) air. This insight steers us towards contributing an attempt at getting further command of the air pollution problem through a PoC of smog modelling with LSTM and GRU models. Specifically, the combined modelling of contaminants NO\tsub{2}, O\tsub{3}, PM\tsub{2.5}, and PM\tsub{10} is considered. 

Ultimately, this research addresses the question: ``To what extent are models with the LSTM and GRU architecture capable of the multivariate timeseries forecasting of smog-related air components?" It is found that the LSTM and GRU can indeed accurately forecast smog-related air components, thus providing an effective pollutant modelling and forecasting method.

\section{Background}
\label{sec:theoretical_framework}
This section briefly explores atmospheric interactions in air pollution and recurrent neural nets.

\subsection{Atmospheric interactions}\label{sec:air_pollution}

The very reality of contaminant concentrations changing over time naturally focuses attention on the question of how these changes originate and evolve.
Whereas the origins and sources of pollution are reasonably well understood \citep{vallero2014fundamentals, saxena2018air}, much is still unknown about its dynamics and how it evolves---hence, the subject of this research.
Especially from a ML perspective, understanding the specific relationships between variables, or features, is critical for efficient learning \citep{li2017feature}.

How pollutants evolve is partly explicable by their interaction with their environment.
As a result, a combined modelling of atmospheric variables can be justified.
The following paragraphs briefly introduce the pollutants' interconnectivity and atmospheric interaction.

Foremost, the chemical interrelation of NO\tsub{2} and O\tsub{3}. Nitrogen oxides (NO\tsub{x}) are (mostly anthropogenically-generated) primary pollutants. When in the presence of certain volatile organic compounds (VOCs) or another initiator or catalyst, NO can oxidate into NO\tsub{2} \citep{atkinson2000atmospheric}.
The photodecomposition of NO\tsub{2} will, in turn, initiate the sequential formation of secondary pollutant O\tsub{3} \citep{finlayson1993atmospheric}.
It follows that, by exclusive consideration and assumption of the presence of photons with adequate energy, NO\tsub{2} affects the atmospheric quantity of O\tsub{3} positively.

When no (or less) solar energy is present (e.g., at night), NO\tsub{2} remains stable and reacts with O\tsub{3} to form nitrate radical (NO\tsub{3}) \citep{finlayson1993atmospheric}, lowering the concentrations of NO\tsub{2} and O\tsub{3} (at least for now).
Therefore, owing to the chemical interdependence of O\tsub{3} and NO\tsub{x}, the levels of O\tsub{3} and NO\tsub{2} are inextricably linked \citep{clapp2001analysis}. Besides, we already observe that one cannot dissociate pollutant concentrations from atmospheric (and cosmic) influences.

Continuing, PM is either emitted directly into the atmosphere (primary) or formed later (secondary) and is subject to air transport, cloud processing, and removal from the atmosphere \citep{seinfeld2016atmospheric}. PM\tsub{10} and PM\tsub{2.5} are interconnected as seen empirically \citep{velders2014grootschalige} and naturally \citep{rhodes2024introduction}, given that their distinction is their size. With its relatively large size, PM itself experiences negligible chemical reactivity with the atmosphere compared to minor compounds such as NO\tsub{2} and O\tsub{3}. Nonetheless, noting its susceptibility to transport, PM and other pollutants alike are responsive to airflow and dispersion in their ambient air environment---an environment amidst all meteorological influences, without yet considering factors such as geology and topology. Therefore, many, at least implicit, parameters are required to model PM and other air components reliably.

In short, NO\tsub{2}, O\tsub{3}, PM\tsub{10}, and PM\tsub{2.5} are subject to influences from all dimensions and thus can be broadly modelled: for modelling pollution movements, pollution can be assumed to behave as air. Furthermore, pollutants are ``internally" affected by each other and externally by the atmosphere, warranting a multivariate, integrated modelling approach.

\subsection{Recurrent Neural Networks}\label{sec:recurrent_neural_nets}

Recurrent neural networks (RNNs) host cyclical connection pathways that, mathematically speaking, represent not functions but dynamical systems (DSs).
RNNs have a network state $\textrm{\bf{x}}(n)$ allowing some earlier input $\textrm{\bf{u}}(n')$ to leave its traces on output $\textrm{\bf{y}}(n)$, and, therewith, the current state is influenced by past states and input.
In practical terms, RNNs are tailored for sequential, fixed order data, such as timeseries.
Illustratively, when data is fed non-sequentially, e.g. today's weather prior to yesterday's, the internal state becomes confounded.

Formally, the transition of an RNN network state
is given by the update equations:
\begin{align}
    \textrm{\bf{x}}(n) &= \sigma(W \textrm{\bf{x}}(n - 1) + W^{\textrm{in}}\textrm{\bf{u}}(n), \label{eq:RNN_state_transition}\\
    \textrm{\bf{y}}(n) &= f(W^{\textrm{out}}\textrm{\bf{x}}(n)),\label{eq:RNN_state_output}
\end{align}
where $n = 0, 1, 2, ..., n_{\textrm{max}}$ are the time steps, $W$ is a matrix containing the connections weights, $W^{\textrm{in}}$ and $W^{\textrm{out}}$ contain the weights from/to the input/output neurons, $\sigma$ is a sigmoid function, and $f$ a function wrapping the readout $W^{\textrm{out}}\textrm{\bf{x}}(n)$ \citep{LN_NN_RUG}.
In particular the activation function $\sigma$, which introduces non-linearity to the evolution of the internal state (\ref{eq:RNN_state_transition}), enables RNNs to capture (long-term) non-linear dependencies in the data.
RNNs are trained with a technique called backpropagation through time (BPTT), which is adapted to their sequential nature.
However, BPTT suffers from vanishing gradients \citep{hochreiter1998vanishing}, making it challenging to learn long-term dependencies effectively \citep{bengio1994learning}.


Long short-term memory (LSTM) networks were introduced by Hochreiter, Schmidhuber, and Gers with the intention to solve this problem \citep{hochreiter1991untersuchungen, hochreiter1997long, gers2000learning}.
They proposed a self-connected linear unit, the LSTM \emph{memory cell}, with a constant error flow: in the absence of new input or error signals to the cell, the local error backflow remains constant, neither growing nor decaying \citep{gers2000learning}. Thus, with the LSTM, the gradient is independent of time.

A more recently proposed recurrent unit is the gated recurrent unit (GRU) by \citet{cho2014properties}.
The GRU uses a similar approach to solving the vanishing gradient problem but simpler.
It contains only two gates, the reset gate and update gate, making it easier to compute.
The former controls the degree to which the previous hidden state influences the current, and the latter combines the LSTM input and forget gate into one.
Its performance has shown to be on par with the LSTM, and, in some cases, can outperform it in terms of convergence in CPU time and in terms of parameter updates and generalisation \citep{chung2014empirical}.

As seen in Section~\ref{sec:state_of_the_art}, the gating mechanism also proves itself in air pollution-related applications \citep{freeman2018forecasting, tao2019air}.
Still, these studies predicted one contaminant only, while LSTMs are proven to be adequate for multivariate data \citep{che2018recurrent}.

At a higher level, beyond the configurations of individual gates or neurons, is where discussion of multivariate data can begin.
A way to strike a balance between a multivariate forecast and an individual one is through non-homogeneous hierarchical neural circuits and architectures, also called hierarchical neural nets (HNNs).
HNNs consist of a number of loosely-coupled subnets, arranged in layers, where each subnet is intended to capture specific aspects of the input data \citep{mavrovouniotis1992hierarchical}.
Such a balance lends itself particularly well to air pollution data, as will be discussed in Section~\ref{sec:model_architecture}.

\vspace{0.65\baselineskip}

\section{Methods}
\label{sec:methods}

The ensuing sections describe the data, preprocessing, model types, training process, and evaluation metrics.

\subsection{Data}\label{sec:data}

The proposed forecasting experiment uses hourly-sampled data from 2016 to 2023 \citep{rivm_rivm_2024, knmi_knmi_2024}, which is available under an initiative of the Dutch government and the Dutch national meteorological service, the Royal Netherlands Meteorological Institute (KNMI).
The data is accreditated under NEN-EN-ISO/IEC 17025 standards and is technically and substantively validated (and possibly rejected) before release \citep{knmi_readme_pdf}.


\begin{figure}[t]
  \centering
  \includegraphics[width = 0.99\columnwidth]{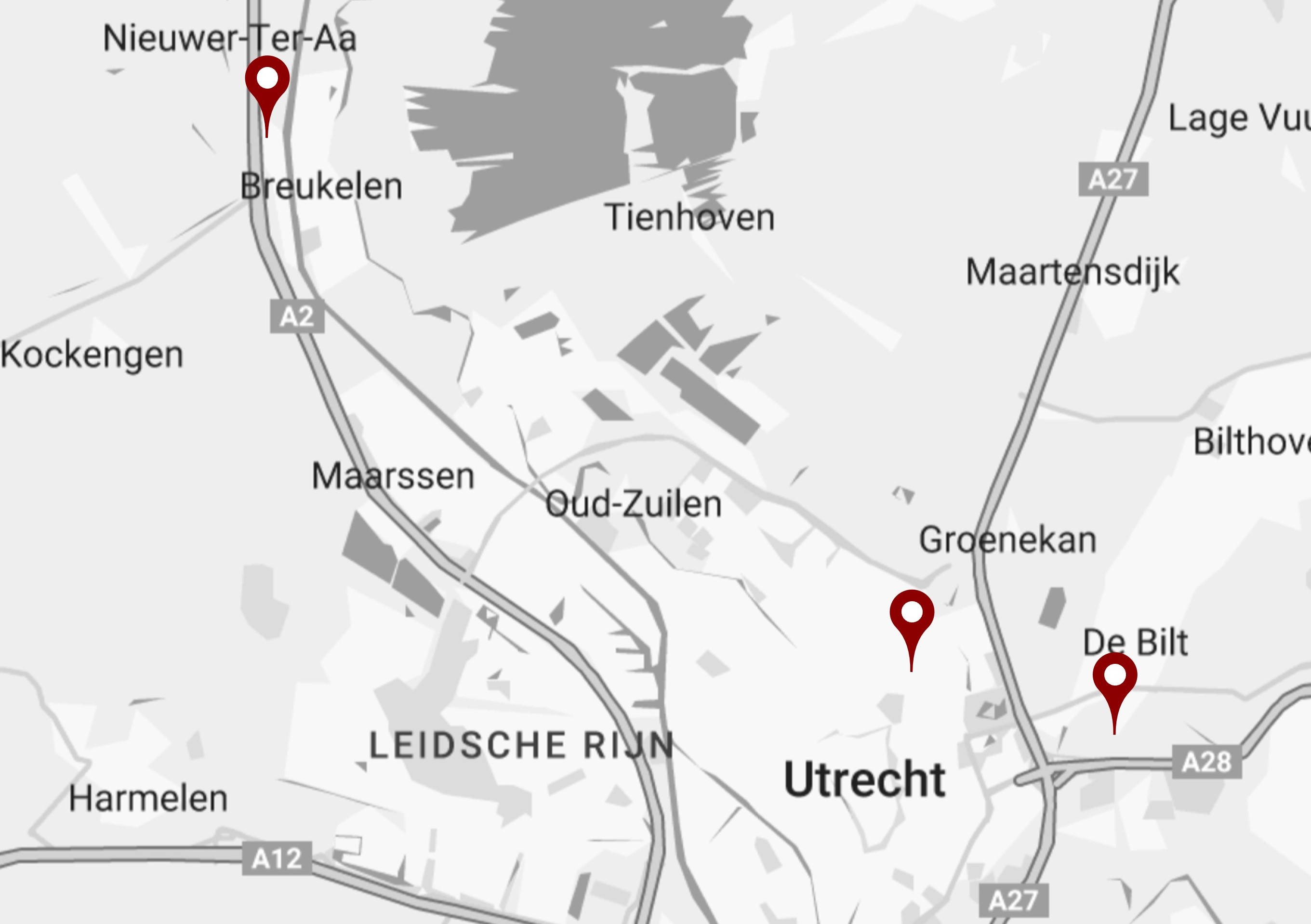}
  \caption{Utrecht area with markers indicating the AWS locations.}
  \label{fig:map_used_sensors}
  \vspace{-0.8\baselineskip}
\end{figure}

\subsubsection{Spatiotemporal context}\label{sec:spatiotemporal_context}

This experiment involves forecasting with data from two locations, a source location ($A$) and a target location ($B$).
The source location is in Utrecht, the Netherlands, and here, pollutant data is combined with meteorologically related covariates from a sensor located in De Bilt, to forecast pollutant data at the relatively northwestern target location in Breukelen. Their relative positions are best illustrated in Figure~\ref{fig:map_used_sensors}.

\subsubsection{Inspection}\label{sec:inspection}

Table~\ref{table:all_paramaters} details the predictive variables used in this experiment, along with the initially considered meteorological parameters.
The rationale and relevance of the meteorological parameters in the context of pollution prediction are discussed in Appendix~\ref{ap:meteo_data_exploration}, coupled with an inspection.

\subsection{Preprocessing}\label{sec:preprocessing}

Preprocessing starts with tidying the raw data, followed by a train-validation-test split, feature engineering, normalisation, and ends with generating (input, output)-pairs.

Firstly, the raw data was cleaned to make it utilisable, for example by solving erronous (split) rows and columns, converting encodings, extracting metadata, and excluding data disqualified due to outliers. Next, there were missing values (Table~\ref{table:missing_values}). These were assumed to be missing completely at random (MCAR) and imputed by linear interpolation.



\begin{table}[ttt] 
\vspace{-0.8\baselineskip}
    \centering
    \caption{Predictive variables and initially considered meteorological variables (in alphabetic order). Some units are multiplied by 0.1 for simplification without losing significance.}
    \vspace{-0.7\baselineskip}
    \label{table:all_paramaters}
    \begin{tabular}[t]{@{}p{0.65\linewidth}p{0.2\linewidth}@{}}
        \toprule
        \hspace{0.25cm}\raggedright\emph{Variable} & \emph{Unit}\hspace{0.25cm} \\
        \midrule
        \hspace{0.25cm}Nitrogen dioxide (NO\tsub{2}) & \SI{}{\micro\gram\per\meter\cubed} \\
        \hspace{0.25cm}Ozone (O\tsub{3}) & \SI{}{\micro\gram\per\meter\cubed} \\
        \hspace{0.25cm}PM $\le$ \SI{10}{\micro\meter} (PM\tsub{10}) & \SI{}{\micro\gram\per\meter\cubed} \\
        \hspace{0.25cm}PM $\le$ \SI{2.5}{\micro\meter} (PM\tsub{2.5}) & \SI{}{\micro\gram\per\meter\cubed} \\
        \midrule
        \hspace{0.25cm}Air pressure (AP) & \SI{0.1}{\hecto\pascal} \\
        \hspace{0.25cm}Dew point temperature (DPT) & \SI{0.1}{\celsius} \\
        \hspace{0.25cm}Global radiation (GR) & \SI{}{\joule\per\square\centi\meter} \\
        \hspace{0.25cm}Maximum wind gust (MWG) & \SI{0.1}{\meter\per\second} \\
        \hspace{0.25cm}Mean wind direction (MWD) & $0-360\degree$ \\
        \hspace{0.25cm}Mean wind speed (MWS) & \SI{0.1}{\meter\per\second} \\
        \hspace{0.25cm}Precipitation amount (PA) & \SI{0.1}{\mm} \\
        \hspace{0.25cm}Precipitation duration (PD) & \SI{0.1}{\hour} \\
        \hspace{0.25cm}Sunshine duration (SD) & \SI{0.1}{\hour} \\
        \hspace{0.25cm}Temperature (T) & \SI{0.1}{\celsius} \\
        \bottomrule
    \end{tabular}
    \vspace{-0.8\baselineskip}
\end{table}

\begin{figure}[bbb]
\vspace{-1.2\baselineskip}
  \centering
  \includegraphics[width = 0.85\columnwidth]{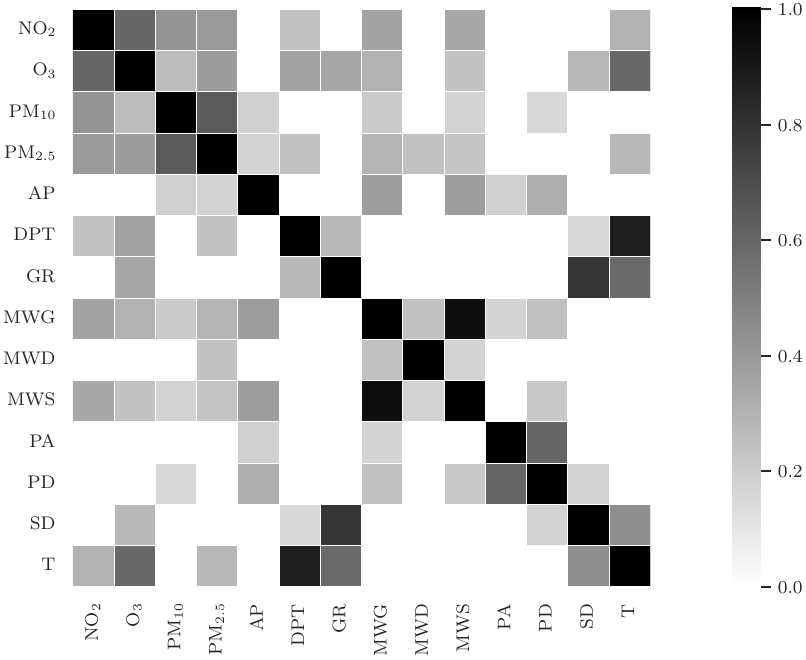}
  \caption{Coefficient matrix for the initially considered features. A threshold $r_{th}$ for the absolute Pearson coefficient is set at $r_{th} = 0.15$. When not met, the entry remains white.
  }
  \label{fig:correlationmatrix}
  \vspace{-0.8\baselineskip}
\end{figure}

Secondly, the tidy data is split into a training, validation, and testing set. This resulted in a balance of $76.3\%, \, 11.9\%, \,11.9\%$.
Subsequently, the newly acquired training set undergoes feature selection.
As described in \citet{hall1999correlation}, good feature sets contain features that are highly correlated with the class, yet uncorrelated with each other.
Thus, to assess the features their intercorrelations are assessed and compared to a threshold $r_{\textrm{th}}$ using the absolute value Pearson correlation coefficient $r_{xy}$:
\begin{equation}\label{eq:pearson_correlation_coefficient}
    r_{xy} =
    \left|\frac
    {\sum^{n}_{i=1}(x_{i} - \bar{x})(y_{i} - \bar{y})}
    {\sqrt{\sum^{n}_{i=1} (x_{i} - \bar{x})^{2}}\sqrt{\sum^{n}_{i=1} (y_{i} - \bar{y})^{2}}}
    \right|,
\end{equation}
where, given paired data $(x_{i}, y_{i})^{n}_{i=1}$, $n$ is the sample size, and $\bar{x}$, $\bar{y}$ are their sample means.
It must be noted, however, that the calculation assumes linear relationships, heteroskedasticity, and a Gaussian distribution.
The correlations are plotted in Figure~\ref{fig:correlationmatrix}.

Upon filtering, the features precipitation amount and duration, global
radiation, and maximum wind gust are discarded, leaving 10 available features.
For details on factors affecting the data split and filtering choices, refer to the source document \citep{oldenburg2024smogclouds}.



Next, normalisation. Normalising promotes generalisation, stabilises gradients and the learning process, and can produce faster convergence \citep{ioffe2015batch}. The selected features are normalised to a range of $\left[0, 1\right]$ with min-maxscaling
\begin{equation}\label{eq:minmax_scaling}
    x' = \frac{x - x_{min}}{x_{max} - x_{min}},
\end{equation}
where $x_{min}$ and $x_{max}$ are the minimum and maximum value for each feature in the training set.

Lastly in preparing the data for the models, pair generation. In order to obtain static input-output pairs from a discretised hourly-sampled temporal training sequence, one segments the input timeseries $(\textrm{\bf{u}}(n))_{n\in[0,N_{u}]}$ for Utrecht, and input timeseries $(\textrm{\bf{y}}(n))_{n\in[0,N_{y}]}$ for Breukelen with $N_{u} = N_{y}$, into sliding windows of length $l_{\textrm{in}} = 72$ and $h = 24$, respectively, obtaining input-output pairs $(\textrm{\bf{u}}_{i}, \textrm{\bf{y}}_{i})_{i = 1,...,P}$ consisting of input
\begin{equation}
    \textrm{\bf{u}}_{i} = (\textrm{\bf{u}}(n_{i}), ..., (\textrm{\bf{u}}(n_{i} + l_{\textrm{in}})),
\end{equation}
and output
\begin{equation}
    \textrm{\bf{y}}_{i} = (\textrm{\bf{y}}(n_{i} + \delta), ...,(\textrm{\bf{y}}(n_{i} + \delta + h)),
\end{equation}
where $n_{i}$ represents the starting index of the $i$-th pair, $P$ denotes the number of pairs as defined by $P = \floor*{\frac{N_{u} + 1}{\Delta n}}$ with sampling step size $\Delta n$, and $\delta$ is defined as $\delta = l_{in} - 24 + 1$, meaning $\textrm{\bf{y}}_{i}$'s output is considered from the 48th hour on, plus a 1-hour window for the spatial prediction from $\textrm{\bf{u}}_{i}$ to $\textrm{\bf{y}}_{i}$.
To expand on the latter, and as seen in \eqref{eq:RNN_state_transition} and \eqref{eq:RNN_state_output}, RNNs process values one-by-one, which for this case means for $l_{in}$ iterations---$\delta$, however, selects only the last 24 readouts for predicting and, thus, learning (facilitated by the loss function~\eqref{eq:loss_function_MSE} discussed in Section~\ref{sec:training}). Step size $\Delta n$ is set at $\Delta n = 24$ for computational efficiency, and because trial-and-error testing showed no or minor upside to a smaller $\Delta n$. 

Essentially, this means that for each pair, an hour of pollutant concentrations at $B$ will be predicted 24 times in sequence, with the preceding hours of $A$ available as the ground for prediction. Thus, the data sets the models up to learn to model the pollutants using their covariates for the spatial prediction task from Utrecht to Breukelen.

\subsection{Model architecture}\label{sec:model_architecture}

The multivariate one-dimensional forecasting task of smog clouds, i.e., modelling the four pollutants from Utrecht to Breukelen, is taken on using six models: an ordinary multi-layer perceptron (MLP), a hierarchical MLP (HMLP), an LSTM and GRU, and, as main contenders, a hierarchical LSTM (HLSTM) and GRU (HGRU).
This section will outline the modelling types and set-ups, followed by their hyperparameter optimization procedure.

\subsubsection{Types}\label{sec:model_types}

As touched upon in Section~\ref{sec:recurrent_neural_nets}, the MLP models approximate not DSs but functions.
Where RNNs have a state $\textrm{\bf{x}}(n)$ allowing some earlier input $\textrm{\bf{u}}(n')$ to leave its traces on output $\textrm{\bf{y}}(n)$, MLPs learn to approximate a (nonlinear) function $f~:~\mathbb{R}^{L^{0}}~\rightarrow~\mathbb{R}^{L^{k}}$, where $L^{0}$ and $L^{k}$ represent the neurons in the input- and output layer, and lack an explicit mechanism for retaining sequences over extended periods.
In practice, they cannot utilise the sequence-spanning BPTT; they propagate errors solely through the network.
Hence, the MLP and HMLP are less suited for this task than RNNs and serve as benchmarks. 


In terms of their specific architecture, the input and output layer are of size $L^{0} = 10$ (ten features) and $L^{k} = 4$ (four predictive variables).
Unidirectional layers knit these together.
For the MLP, these are standard fully-connected layers.
Its counterpart, the HMLP, is of the type of hierarchical models---a term introduced in Section~\ref{sec:recurrent_neural_nets} which describes non-homogeneous, modular neural circuits. HNNs can, depending on the task, perform multi-task learning (MTL), a method whose principle goal is to improve generalisation performance \citep{caruana1997multitask}.

Furthermore, with HNNs, the hierarchical organisational structure is in hands of the model designer and offers an opening for a priori knowledge to be embodied in the neuronal arrangement as inductive bias or regularising factor, guiding the model in a preferred direction.
In the context of this study, we aim to predict the four pollutants, each of which can be regarded as a distinct subtask.
Recognising both the intercorrelations of the pollutants (as depicted, for example, in Figure~\ref{fig:correlationmatrix}) and the fact that they all live a life of their own, it seems reasonable to mirror this reality in a model's architecture.
To achieve this, we employ one shared layer to establish shared representation and subsequently partition the network flow into a modular branch per subtask to reduce the interference between tasks.
This design, including this nuanced regularising factor, confers HNNs a hypothetical advantage over fully-connected nets, which neglect an explicit internal-external balance. 

Next are the RNNs. The RNNs use the PyTorch implementation of LSTM and GRU memory cells introduced in Section~\ref{sec:recurrent_neural_nets}.
The fully-connected RNNs are similar to the MLP, except for their gating mechanisms and recurrent synaptic connections, and vice versa for the hierarchical RNNs in relation to the HMLP.

Following up on identifying model types, hyperparameters reveal in more detail how these types are shaped into complete architectures.
The following sections discuss how they are established.



\vspace{0.65\baselineskip}
\subsubsection{Hyperparameter optimization}\label{sec:hyp_optimization}

Hyperparameters can be used to control the behaviour of a learning algorithm and are not adapted by the algorithm itself \citep{goodfellow2016deep}.
For the DL models at hand, examples are the number of hidden layers and hidden units, the learning rate, choice of optimizer, and the regularisation term.
An overview of the used hyperparameters per model can be found in Appendix~\ref{ap:architecture_details}---this section will chiefly explore the methodology behind their selection.

To determine their values, a hyperparameter search procedure consisting of a grid search and cross-validation (CV) schemes is used.
This procedure aims to find a hyperparameter configuration $c$ with a minimised loss, while also testing $c$'s generalisation capabilities.
The loss is calculated on distinct validation sets generated by CV from all available training data, i.e. a concatenation of the training and validation set, to test this generalisation performance and prevent overfitting.
Nested within the hyperparameter search and within the CV scheme, is the models' training algorithm, which, together with the loss, is specified in the next subsection, Section~\ref{sec:training}.

Continuing, the traditional grid search was used because of its ease suiting this PoC; it essentially does a brute-force search through the parameter space $H$.
Here, $H$ is defined as the Cartesian product of the finite sets $S$ containing possible values for each parameter.
Because $H$ grows exponentially, a large $S$ is not feasible, and smaller $S$ are already computationally expensive.
Hence, as measures, some initial trial runs were executed to get a feel of which options to include and the many models were computed on an HPC cluster.


Then, CV is run for each $c$, where $c$ is a unique configuration within $H$.
For the stateless MLPs, regular $k$-fold cross-validation, with $k = 5$, is used---with the perk of maximal data usage.
RNNs, conversely, do have a state and allow memory trace of past sequences, as aforementioned in Section~\ref{sec:recurrent_neural_nets}.
A variation of $k$-fold CV, called sliding window CV, accommodates this: it samples training and validation sets---with, in contrast to pair generation, superposition of intervals---using a sliding window approach, thus not allowing validation of trained models with out-of-sample data directly from the past.

With these schemes, grid search with $k$-fold CV for the MLPs and sliding-window CV for the RNNs, values for the hyperparameters were determined---forging the model types into architectures.



\vspace{0.2\baselineskip}
\subsection{Training}\label{sec:training}

In this section, we explain the training procedure used during hyperparameter optimization and the final training itself by defining the optimization goal and method, followed by some anti-overfitting measures.
The final models are trained using the training and validation set created in Section~\ref{sec:preprocessing}.


To approximate a model $m_{\theta}$ parametrised by tuneable parameters collected in a vector $\theta$, given a search space $\theta \in \Theta$ of target models $\Theta$ within the same architecture; training pairs $(\textrm{\bf{u}}_{i}, \textrm{\bf{y}}_{i})_{i = 1,...,P}$; and the mean squared error (MSE) loss function defined as
\begin{equation}\label{eq:loss_function_MSE}
    \text{MSE} = \frac{1}{n} \sum_{i=1}^{n} (y_i - \hat{y}_i)^2 + \lambda \mid \mid \theta \mid \mid_2^2,
\end{equation}
where $n$ denotes sample amount, $y$ the ground truth, and $\hat{y}$ the prediction---one has to solve the optimization problem
\begin{equation}
    \theta_{\textrm{opt}} =
    \underset{\theta \in \Theta}{\textrm{argmin}}\frac{1}{P}\sum_{i=1,...,P}\text{MSE}(m_{\theta}(\textrm\bf{x}_{i}), \textrm\bf{y}_{i}),
\end{equation}
where $\theta_{\textrm{opt}}$ denotes a model with a minimised empirical risk \citep{LN_ML_RUG, LN_NN_RUG}.
MSE punishes extreme values quadratically more, suiting the context of air pollution where extremes are of greater concern.


With an initial model $\theta^{(0)}$, initialised by the PyTorch-default Kaiming initialisation \citep{he2015delving}, optimization of $\theta^{(n)}$ is performed by the \emph{Adam} (ADAptive Momentum) optimizer \citep{kingma2014adam}.
Adam differentiates itself from, e.g., stochastic gradient descent (SGD) by using momentum \citep{sutskever2013importance} and adaptive learning rates per parameter while only requiring first-order gradients and little memory \citep{kingma2014adam}.
This study uses its implementation in PyTorch.


Adam does its work every time a batch $B$ of 16 $(\textrm{\bf{u}}, \textrm{\bf{y}})$-pairs is passed.
Batches are randomly sampled (while in sequence order) from the available pairs, introducing stochasticity (and efficiency over, e.g., one-by-one calculation).
Internally, this adds the batch dimension to the pairs, creating the tensors [$B$,~$l_{in}$,~$L^{0}$] for $\textrm{\bf{u}}$ and [$B$,~$h$,~$L^{k}$] for $\textrm{\bf{y}}$.
When $\textrm{\bf{u}}$ is fed, the models spit out forecasts $\textrm{\bf{y}}'$ in the form of such tensor, which is subsequently compared to the ground truth $\textrm{\bf{y}}$ yielding the loss with which $\theta$ can be updated.


Updating $\theta$, however, proceeds quite differently for the two main types of models.
Whilst for the fully-connected models, this proceeds as usual with one optimizer updating $\theta$, the modular models require a different approach.
As they essentially consist of multiple core components (one shared layer, four branches) with different search spaces and convergence qualities, the process capitalises on this: all five components have their own optimizer and matched scheduler.
In addition, they have two separate (initial) learning rates, as seen in Table~\ref{table:hyperparams} and Table~\ref{table:hyperparams_complete}.
Distributing the learning tasks helps each model part stably reach an optimum.

Lighting this in terms of implementation, the shared and branched parts do epochs in turns, seeing all the batches separately while the other is \emph{frozen}.
Frozen, as in, the parameters cannot update but can infer.
A con here is efficiency: the batches are passed through the model once more for every epoch.

When zooming out and looking at when learning should finalise, early stopping comes in: it finishes training when for some number of epochs (defined in Table~\ref{table:training_settings_complete}) the validation loss does not decrease by $\geq~\num{1e-5}$.
Another anti-overfitting measure is the $L2$-norm regularisation added to \eqref{eq:loss_function_MSE}.
As effect, larger weights are penalised and smaller weights are encouraged, preventing some set of weights dominating the model.


In summary, the training process seeks to find an optimal set of model weights $\theta_{opt}$, and to regularise the learning process early stops, the batches introduce stochasticity, the regularisation term balances weight values, and the hierarchical nets incorporate MTL. With these regularisation steps, the training procedure should yield models fulfilling the ultimate objective of generalisation, tested in next section.






\subsection{Evaluation}\label{sec:evaluation}

For evaluation of the models, the held-out test set is used.
The predictions were first post-processed using inverse minmax-scaling, sampled in batches without shuffling to eliminate any randomness, and then evaluated using the root mean squared error (RMSE) and symmetric mean percentage error (sMAPE) metric, which both serve a different interpretation of model performance. In addition, the inference speed of each model is evaluated, as this is one of the unique advantages of data-driven methods over first-principle methods like NWP.

The RMSE, defined as
\begin{equation}
    \text{RMSE} = \sqrt{\frac{1}{n} \sum_{i=1}^{n} (y_i - \hat{y}_i)^2},
\end{equation}
provides a measure of the average magnitude of the error with, due to its squaring operation, larger errors getting penalised more. This fits the context of smog modelling, where higher values are especially harmful. 

The other metric, the sMAPE:
\begin{equation}\label{eq:sMAPE}
    \text{sMAPE} = \frac{2}{n} \sum_{i=1}^{n} \frac{\left| y_i - \hat{y}_i \right|}{ \left( \left| y_i \right| + \left| \hat{y}_i \right| \right)} \times 100,
\end{equation}
is an accuracy measure based on percentage (or relative) errors, providing a scale-independent, well-interpretable metric. The sMAPE complements the RMSE by taking into account the individual and different distributions of the pollutants. Therefore, the metric allows for a fair relative comparison of the models' performance.

Finally, it is worth emphasising that, generally speaking, the RMSE serves a practical purpose because it tells about the deviation in \SI{}{\micro\gram\per\meter\cubed} and has a quadratically progressive penalty. The sMAPE mainly fulfils a ``scientific" purpose due to the possibility of comparing models, though the two metrics are not mutually exclusive.
This idea acts as a guide to interpret the results meaningfully.

\section{Results}
\label{sec:results}
Following training, the models are evaluated on out-of-sample data.
It is found that the models provide an effective method for the modelling and forecasting of the pollutants.
Quantitative results by RMSE and sMAPE are listed in Table~\ref{table:result_metrics}.

Considering the subtask-specific lowest sMAPE values, NO\tsub{2} is predicted most accurately.
Following NO\tsub{2} is O\tsub{3}, then PM\tsub{2.5}, and the models were least successful in predicting PM\tsub{10}.
Nonetheless, the lowest sMAPEs, as well as the RMSEs---\textbf{which are primarily generated by the HGRU}---confirm the models' suitability for forecasting the pollutants at $B$ using data at $A$.
Meanwhile, the models also differed in performance.

Measured by RMSE, the non-hierarchical fully-connected RNNs perform predominantly better than the MLPs, but also utilise many parameters to do so.
Measured by sMAPE, they do too, despite HMLP's sMAPE $(M=46.274,SD=45.344)$ being slightly inferior to the LSTM's $(M=46.321,SD=46.515)$: 
a paired t-test with $\alpha < .05$ suggests there is no sufficient evidence to reject the null hypothesis of no difference, $t(8927)=1.011,p=\SI{3.12e-01}{}$.

Furthermore, the GRU yields the lowest errors of the non-hierarchical RNN models.
The HLSTM ranks second, and, as per RMSE and sMAPE, the HGRU performs best, establishing the hierarchical models as the top performers.
A paired t-test confirms the HGRU's ($M_{\textrm{RMSE}}=5.468,SD_{\textrm{RMSE}}=4.906$,$M_{\textrm{sMAPE}}=44.519,SD_{\textrm{sMAPE}}=44.519$) significant predictive ability on the testing set over the HLSTM ($M_{\textrm{RMSE}}=5.633,SD_{\textrm{RMSE}}=4.935$,$M_{\textrm{sMAPE}}=44.981,SD_{\textrm{sMAPE}}=45.850$) both by RMSE ($t(8927)=-5.922,p=\SI{3.30e-9}{}$) and sMAPE ($t(8927)=-2.855,p=\SI{4.32e-3}{}$), as well as on the other models. Moreover, for all individual pollutant subtasks by RMSE, and most by sMAPE, the HGRU exhibits the highest predictive precision, where it is only surpassed repeatedly with the sMAPE of the PM\tsub{2.5}-subtask.

\begin{table*}[tt] 
\sisetup{detect-weight=true,detect-inline-weight=math}
    \centering
    \caption{Results of each model, evaluated and compared on performance (RMSE and sMAPE) and efficiency (inference speed $t_\textrm{\textbf{inf}}$ and parameter count). The error metrics are calculated per pollutant and combined, with the lowest error in \textbf{bold}. $t_\textrm{\textbf{inf}}$ is the time in milliseconds for one inference of one 24-hour lead time prediction (processed on an Intel Core i7-8565U CPU, 8GB RAM, 64-bit OS).}
    \vspace{0.1\baselineskip}
    \label{table:result_metrics}
    \begin{tabular}{l@{\hspace{0.3cm}}
    S[table-format=2.2]@{\hspace{0.11cm}}S[table-format=2.2]@{\hspace{0.11cm}}S[table-format=2.2]@{\hspace{0.11cm}}S[table-format=2.2]@{\hspace{0.11cm}}S[table-format=2.2]
    S[table-format=2.2]@{\hspace{0.14cm}}S[table-format=2.2]@{\hspace{0.14cm}}S[table-format=2.2]@{\hspace{0.14cm}}S[table-format=2.2]@{\hspace{0.14cm}}S[table-format=2.2]
    S[table-format=0.4]@{\hspace{0.18cm}}S[table-format=6.0]
    }
        \toprule
        Models & \multicolumn{10}{c}{Performance} & \multicolumn{2}{c}{Efficiency} \\
        \cmidrule(l){2-11} \cmidrule(l){12-13}
        & \multicolumn{5}{c}{\textbf{RMSE} {(\SI{}{\micro\gram\per\meter\cubed})}} & \multicolumn{5}{c}{\textbf{sMAPE} {(\%)}} & \textbf{$t_\textrm{\textbf{inf}}$}\hspace{0.11cm}{(ms)} & \textbf{Param \#} \\
        \cmidrule(lr){2-6} \cmidrule(lr){7-11}
        & NO\tsub{2} & O\tsub{3} & PM\tsub{10} & PM\tsub{2.5} & Total & NO\tsub{2} & O\tsub{3} & PM\tsub{10} & PM\tsub{2.5} & Total \\
        \addlinespace%
        MLP   & 6.63 & 7.53 & 7.82 & 4.85 & 6.71 & 35.89 & 41.90 & 65.24 & 53.15 & 49.04 & 27.2 & 17604  \\
        HMLP  & 5.99 & 6.83 & 7.95 & 4.62 & 6.35 & 31.84 & 39.44 & 65.42 & \bfseries 48.00 & 46.27 & 135.2 & 15620 \\
        \addlinespace%
        LSTM  & 5.97 & 6.39 & 7.48 & 4.32 & 6.04 & 32.09 & 38.09 & 63.40 & 51.70 & 46.32 & 14.4 & 572640  \\
        HLSTM & 5.36 & 6.57 & 6.60 & 4.00 & 5.63 & \bfseries 28.53 & 38.83 & 60.80 & 51.76 & 44.98 & 18.7 & 72244  \\
        \addlinespace%
        GRU   & 6.01 & 6.18 & 6.84 & 3.94 & 5.74 & 32.62 & 38.46 & 61.15 & 49.67 & 45.47 & 47.9 & 363360  \\
        HGRU  & \bfseries 5.35 & \bfseries 6.01 & \bfseries 6.59 & \bfseries 3.92 & \bfseries 5.47 & 28.78 & \bfseries 36.97 & \bfseries 59.92 & 52.40 & \bfseries 44.52 & 77.4 & 74948  \\
        \bottomrule
    \end{tabular}
\end{table*}

\begin{figure}[t]
  \centering
  \includegraphics[width = 0.48\textwidth]{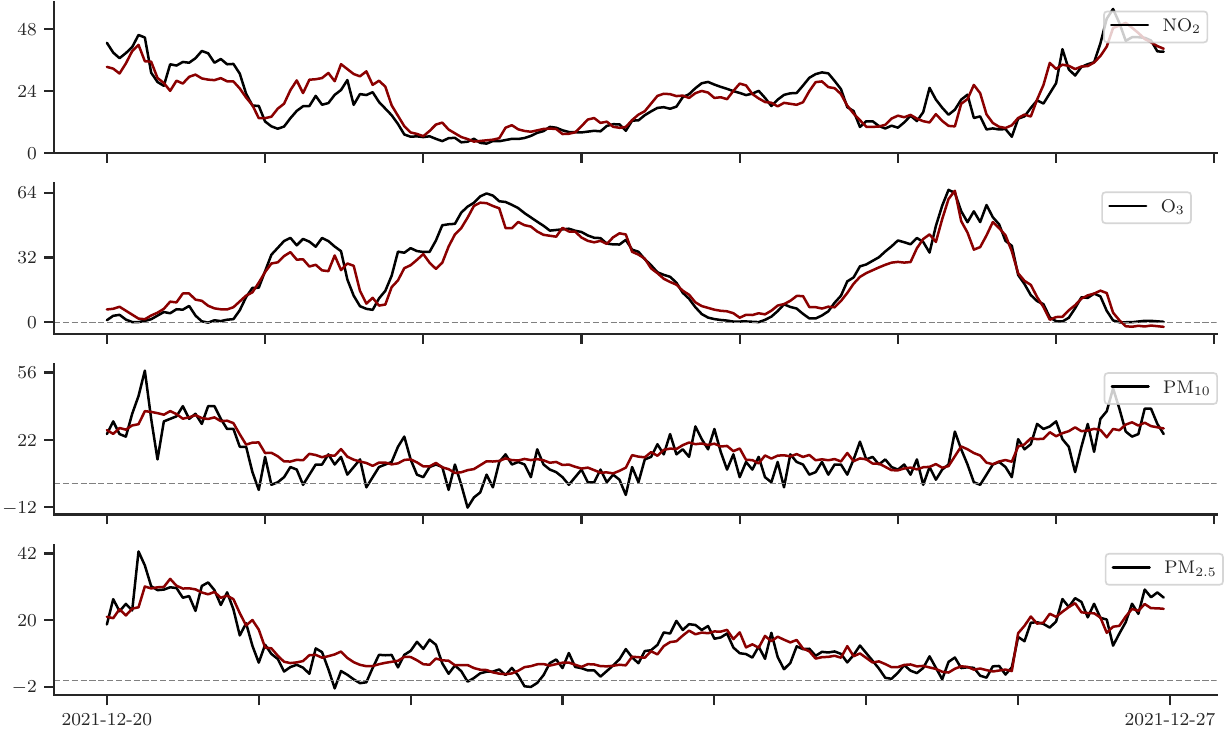}
  \vspace{-1\baselineskip}
  \caption{HGRU forecasts for NO\tsub{2}, O\tsub{3}, PM\tsub{10}, and PM\tsub{2.5} taken for a week from the evaluation set. Black indicates the ground truth and maroon the forecasts. Dashed lines indicate zero.}\label{fig:results_HGRU_small}
  \vspace{-1\baselineskip}
\end{figure}

A line plot, shown in Figure~\ref{fig:results_HGRU_small}, visually represents the HGRU's forecasts.
Consistent with the numerical interpretation of the RMSE and sMAPE, the patterns of NO\tsub{2} and O\tsub{3} seem to be most closely captured.
In contrast, the PMs exhibit more short-term fluctuations that are less frequently captured, with PM\tsub{10} proving the most challenging.
Altogether, it can be stated that the HGRU is well equipped to use data at $A$ for forecasting at $B$.

In terms of efficiency, the inference speed $t_\textrm{inf}$ of the models, as also seen in Table~\ref{table:result_metrics}, shows that efficiency is high: a 24-hour prediction is generated with negligible delay on a relatively inefficient processor.
Counterintuitively, the model with the most parameters is the quickest, though the margins are small.
As also discussed in \citet{pathak2022fourcastnet} and Section~\ref{sec:state_of_the_art} (or, e.g., in the recent \citet{bodnar2024aurora}), the speeds, apart from the initial training cost, highlight the operational advantage of DL models over traditional first-principle methods: they are orders of magnitude faster and more efficient.
Last to note on efficiency is that the best-performing models, the HLSTM and HGRU, require significantly fewer parameters (due to reduced parameter sharing) than the non-hierarchical RNNs as well.

\section{Conclusions}
\label{sec:conclusions}
In this paper, multivariate timeseries forecasting of smog clouds, represented by NO\tsub{2}, O\tsub{3}, PM\tsub{10}, and PM\tsub{2.5} concentrations, using RNNs is conducted.
Specifically, meteorological and pollution data at $A$ is used to forecast air pollution levels at $B$.
The most sophisticated models, the HLSTM and HGRU, are benchmarked with unidirectional and fully-connected DL architectures.

The research question, ``To what extent are models with the LSTM and GRU architecture capable of the multivariate timeseries forecasting of smog-related air components?" is answered by the fact that the models are indeed highly adequate.
Results demonstrate that, above all, the HGRU is suitable and competitive at this task.
Reasons include the sequence-processing prowess of RNNs, a GRU's simplicity, and an integrated design streamlined to the very nature of the pollutants.

To sum up, our study contributes a PoC of smog cloud modelling using RNNs with a hierarchical architecture, providing a basis for advancements in pollution and weather forecasting to improve future public health.
\vspace{-0.21\baselineskip}

\section*{Acknowledgements}
The authors thank the reviewers for their useful comments. Thanks should also go to the Center for Information Technology of the University of Groningen for providing access to the H\'abr\'ok HPC cluster.
\vspace{-0.21\baselineskip}

\section*{Broader Impact Statement}
Air pollution stands as a critical global challenge to humanity \citep{agenda2023}.
The rise of large-scale combustion and anthropogenic polluting activities has led to significant increases in air pollutant concentrations over the last century, leaving a heavy burden on human health \citep{kampa2008human}.
An unmistakable manifestation of these developments is the occurrence of smog: a noxious mixture of air pollutants that obstructs visibility and severely impairs human health in various ways \citep{brunekreef2002air}.
Given the detrimental effects, it is imperative to be able to predict when harmful pollutant levels might occur.
This research proposes different methods to gain insight into air pollutant levels through timeseries forecasting and the application of multiple deep neural architectures, notably including RNNs.

This research's limitations are summarised by simplifying measures to keep it within a scope appropriate for a PoC and by conceptually inherent limitations. Notable inherent limitations of this study include: the data being limited to merely two sensors, which fails to honour the complexity of the modelled system, e.g. the multidimensionality, emission sources, geographical features; the non-stationarity of the data not being explicitly taken into account neither in preprocessing nor in model design; and modelling at a location where the air pollution and smog clouds problem is almost absent, thus limiting the direct impact.

Despite these limitations, recurrent deep architectures offer a promising addition to NWP, given their adequacy and efficiency. Additionally, the minimal data transformation required enables real-time and continuous predictions.

\bibliography{literature}
\bibliographystyle{icml2024}

\newpage
\appendix
\onecolumn
\section{Data insights}\label{ap:data_insights}

This Appendix section provides some additional insight into the data by exploring the meteorogical variables, displaying an overview of the data availabilities, and presenting some data statistics.

\subsection{Exploration of meteorological data}\label{ap:meteo_data_exploration}

This subsection provides an overview of all the initially considered meteorological variables accompanied by an explanation of why these variables might be helpful for modelling smog clouds, i.e. the pollutants NO\tsub{2}, O\tsub{3}, PM\tsub{10}, PM\tsub{2.5}. Important to emphasise is that not all of these rationales have held up in feature selection (or, worded differently, showed in the data).

It is important to stress that besides the individual characteristics of the pollutants, they are located in the tropospheric sky, and have such a low mass they can be assumed to behave as air in terms of their interaction with large-scale meteorological processes. We will go through the variables from top to bottom to explain their relevance:
\begin{itemize}
    \item \textbf{Air pressure} can indicate dispersion and transport of large air currents, for example by low and high-pressure areas, and thus also influence the air currents of pollutants \citep{holton2012introduction}.
    \item \textbf{Dew point temperature, precipitation sum, and precipitation amount} are indicators of atmospheric moisture content levels. These levels can say something about, for example, the rate of condensation (which, via nucleation, can lead to the formation of fine aerosols (PM) \citep{gelbard1979general}), any scavenging and cleansing of the air by rainfall (lowering the pollutant concentrations) \citep{vallero2014fundamentals}, and the formation of acid rain (where acidic gases SO\tsub{2} and NO\tsub{x} that are (related to) the predictive variables, get washed out, thus lowering the concentrations \citep{irwin1988acid}).
    \item \textbf{Global radiation and sunshine}, which signify the presence of solar energy in the form of photons, serve as fundamental drivers of low-entropy energy input on Earth. Section~\ref{sec:air_pollution} discussed an example of a photoinduced process.
    \item \textbf{Temperature} is an essential factor in chemical processes seen by its role as accelerator in the formation of secondary pollutants. In addition, temperature plays a role in atmospheric stability, with, for example, (suddenly) high temperatures signifying increased convective activity. Furthermore, temperature influences state changes, and is also tightly connected with global radiation and sunshine, therewith also indirectly contributing to their effects. For more context on atmospheric chemistry and physics, refer to the extensive \citet{seinfeld2016atmospheric}.
    \item \textbf{Mean wind direction, mean wind speed, and maximum wind gust} all tell about the wind's properties, which in turn carries the pollutants through the atmosphere. In context of the experiment, wind direction, for example, tells about the relative directional relationship between $A$ and $B$. Out of the pollutants, the wind especially plays a role for the PMs, as they have a bigger surface.
\end{itemize}

\subsection{Data availability}\label{ap:data_availability}

Here is a short summary of the availability of the data used in the experiment. Missing data was interpolated with linear interpolation.

\begin{table}[h] 
    \centering
    \caption{Data availability percentage for the modelled pollutants for each year. The meteorological abbrevations are defined in Table~\ref{table:all_paramaters}. The meteorological data was completely available for all years---for the pollutants, it was not. Given the strict procedures by the KNMI \citep{knmi_readme_pdf}, this is no surprise.}
    \label{table:missing_values}
    \begin{tabular}{lllllllllll}
        \toprule
        & NO\tsub{2} & O\tsub{3} & PM\tsub{10} & PM\tsub{2.5} & AP & DP & MWD & MWS & SD & T \\
        \midrule
        2017 & $97.64\%$ & $96.85\%$ & $97.62\%$ & $99.10\%$ & $100\%$ & $100\%$ & $100\%$ & $100\%$ & $100\%$ & $100\%$  \\
        2018 & $99.67\%$ & $98.52\%$ & $97.70\%$ & $99.29\%$ & $100\%$ & $100\%$ & $100\%$ & $100\%$ & $100\%$ & $100\%$  \\
        2020 & $98.60\%$ & $98.05\%$ & $99.34\%$ & $99.31\%$ & $100\%$ & $100\%$ & $100\%$ & $100\%$ & $100\%$ & $100\%$  \\
        2021 & $99.67\%$ & $98.55\%$ & $98.88\%$ & $99.29\%$ & $100\%$ & $100\%$ & $100\%$ & $100\%$ & $100\%$ & $100\%$  \\
        2022 & $96.90\%$ & $97.62\%$ & $95.56\%$ & $99.62\%$ & $100\%$ & $100\%$ & $100\%$ & $100\%$ & $100\%$ & $100\%$  \\
        2023 & $98.60\%$ & $97.15\%$ & $98.46\%$ & $97.97\%$ & $100\%$ & $100\%$ & $100\%$ & $100\%$ & $100\%$ & $100\%$  \\
        \bottomrule
    \end{tabular}
\end{table}

\newpage



\subsection{Data allocation and quantity}\label{ap:data_statistics}

This section provides transparency on how much data was used and in what proportions. Table~\ref{table:data_hours} shows the number of hours of data before pair generation, and Table~\ref{table:data_points} the data after pair generation.

\begin{table}[h] 
    \centering
    \caption{Hours of data for each feature per year in the training, validation, and testing sets before pair generation, illustrating the data balance between the different sets. Divide these by 24 for the amount of days. Abbrevations are defined in Table~\ref{table:all_paramaters}. For further details on the establishment of these balances, refer to the source document \citep{oldenburg2024smogclouds}.
    }
    \label{table:data_hours}
    \begin{tabular}{l@{\hspace{1cm}}*{10}{S[table-format=4.0]}}
        \toprule
        & {NO\tsub{2}} & {O\tsub{3}} & {PM\tsub{10}} & {PM\tsub{2.5}} & {AP} & {DP} & {MWD} & {MWS} & {SD} & {T} \\
        \midrule
        Train '17 & 3648 & 3648 & 3648 & 3648 & 3648 & 3648 & 3648 & 3648 & 3648 & 3648  \\
        Train '18 & 3648 & 3648 & 3648 & 3648 & 3648 & 3648 & 3648 & 3648 & 3648 & 3648  \\
        Train '20 & 3648 & 3648 & 3648 & 3648 & 3648 & 3648 & 3648 & 3648 & 3648 & 3648  \\
        Train '21 & 2640 & 2640 & 2640 & 2640 & 2640 & 2640 & 2640 & 2640 & 2640 & 2640  \\
        Train '22 & 2640 & 2640 & 2640 & 2640 & 2640 & 2640 & 2640 & 2640 & 2640 & 2640  \\
        \midrule
        Validation '21 & 504 & 504 & 504 & 504 & 504 & 504 & 504 & 504 & 504 & 504  \\
        Validation '22 & 504 & 504 & 504 & 504 & 504 & 504 & 504 & 504 & 504 & 504  \\
        Validation '23 & 1512 & 1512 & 1512 & 1512 & 1512 & 1512 & 1512 & 1512 & 1512 & 1512  \\
        \midrule
        Test '21 & 504 & 504 & 504 & 504 & 504 & 504 & 504 & 504 & 504 & 504  \\
        Test '22 & 504 & 504 & 504 & 504 & 504 & 504 & 504 & 504 & 504 & 504  \\
        Test '23 & 1512 & 1512 & 1512 & 1512 & 1512 & 1512 & 1512 & 1512 & 1512 & 1512  \\
        \bottomrule
    \end{tabular}
\end{table}

\begin{table}[h] 
    \centering
    \caption{Table with numerical descriptions of the used datasets, after pair generation performed in Section~\ref{sec:preprocessing} (with $ \Delta n$ = \SI{24}{} hours). Due to the overlapping nature of the pair generation algorithm, ``more" usable data was generated compared to the original data. The amount of pairs $ P$ is displayed, the total amount of hours, total datapoints, datapoints passed through the model as input \textrm{\bf{u}}, and the ground truth \textrm{\bf{y}} datapoints used for the loss function during training, giving an indication of the amount of computations needed for one training epoch. (With the ``optimal" $ \Delta n = \SI{1}{}$, the training set would grow to the impractical amount of \SI{12847104}{} datapoints.)}
    \label{table:data_points}
    \begin{tabular}{lS[table-format=3.0]S[table-format=6.0]S[table-format=6.0]S[table-format=6.0]S[table-format=6.0]}
        \toprule
        & {$P$} & {$hrs_{\textrm{total}}$} & {$n_{\textrm{total}}$} & {$n_{\textrm{\bf{u}}}$} & {$n_{\textrm{\bf{y}}}$} \\
        \midrule
        Training set   & 656 & 47232 & 535296 & 472320 & 62976 \\
        Validation set & 93  & 6696   & 75888   & 66960 & 8928 \\
        Testing set    & 93  & 6696   & 75888   & 66960 & 8928 \\
        \bottomrule
    \end{tabular}
\end{table}

\newpage
\section{Training insights}\label{ap:training_insights}

\begin{figure*}[hhh]
  \centering
  \includegraphics[width = 1\textwidth]{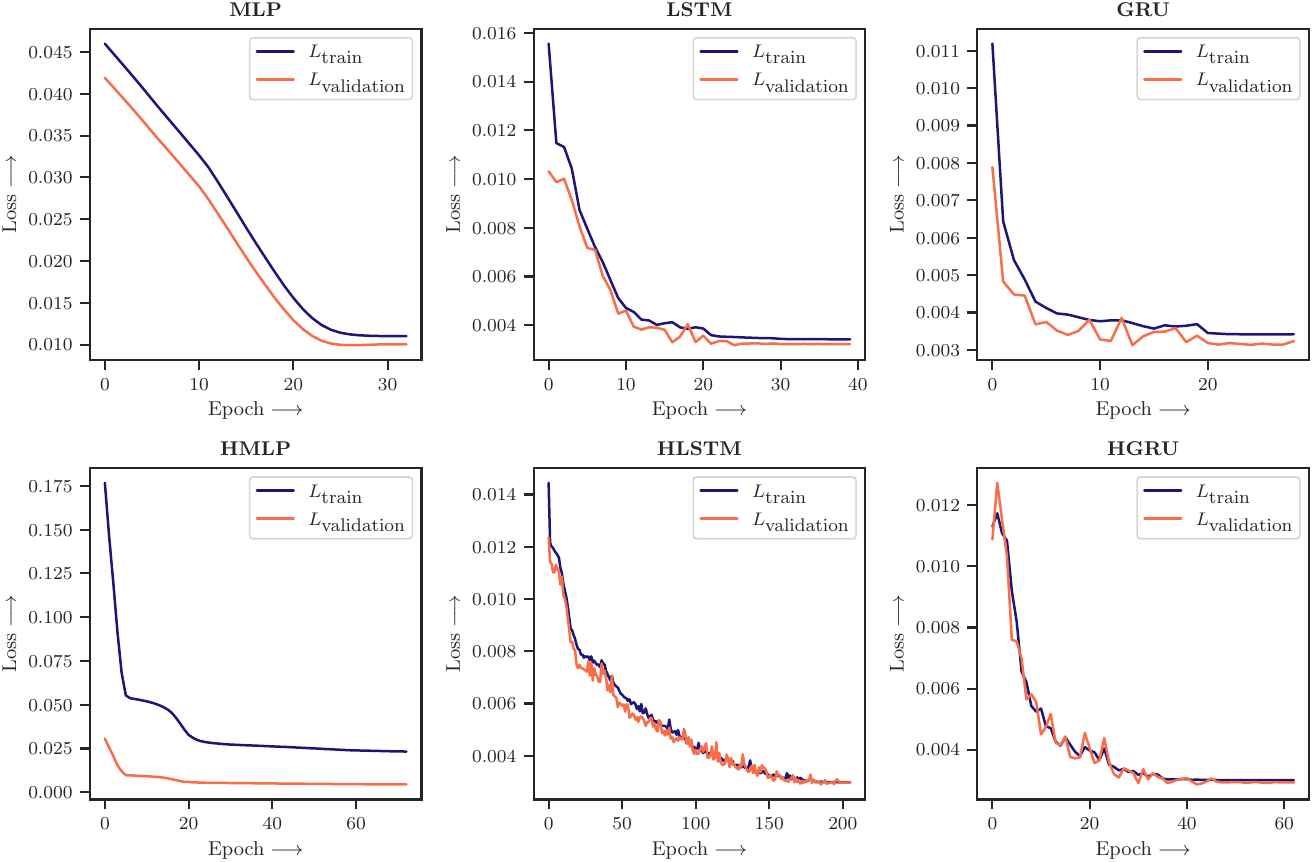}
  \vspace{-1\baselineskip}
  \caption{Loss plots for all models, showing the training versus validation losses over epochs.}\label{fig:losses_all}
\end{figure*}

\begin{figure*}[hhh]
  \centering
  \includegraphics[width = \textwidth]{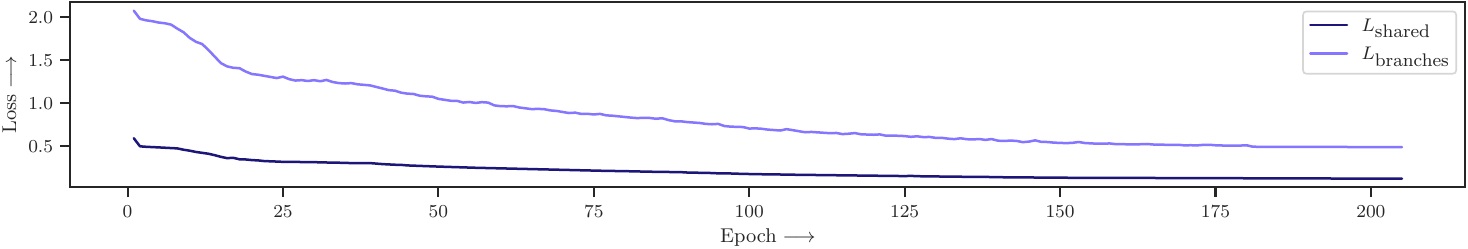}
  \vspace{-1\baselineskip}
  \caption{Training loss plotted of the shared and branched part of the HLSTM. For illustrative purposes, the first epoch is left out from the plot. Both model parts have different complexities (see Section~\ref{sec:model_architecture}), causing their learning process to be different as well. The branches were more complex, causing its learning process to be less stable, visible by the small ``bumps" in its descend.}\label{fig:losses_HLSTM_dist}
  \vspace{-4\baselineskip}
\end{figure*}

The subplots in Figure~\ref{fig:losses_all} show the training and validation loss development during final training of the six models. Figure~\ref{fig:losses_HLSTM_dist} shows how both the shared and branched part of the HLSTM contributed to its training loss.

Training the models took an hour maximum, using the hyperparameters listed in Table~\ref{table:hyperparams_complete} and \ref{table:training_settings_complete} and processed locally on an Intel Core i7-8565U CPU, 8GB RAM, 64-bit OS.

\vspace{2\baselineskip}

\newpage
\section{Architectures}\label{ap:architecture_details}

This Appendix section provides some additional detail on the employed architectures by specifying the used hyperparameters and other model parameters. The used hyperparameter optimization procedure can be found in Section~\ref{sec:hyp_optimization}.

Table~\ref{table:hyperparams} shows the hyperparameters included in the grid search, including their symbols, and Table~\ref{table:hyperparams_complete} shows the determined values as a result of the search.

Some things are worth pointing out here. Note that the fully-connected models have just one learning rate, while the hierarchical models have two: one for their shared layer and one for the optimizers of each of their branches. The ratio between these two, $\mu_{\textrm{shared}}$ and $\mu_{\textrm{branch}}$, is equivalent to $k$. This was done, after lots of test runs, with the idea of a ``power ratio" between the two: the branches needed a higher $\mu$ to let them converge in harmony with the shared layer. Another reason was that by interlinking the two, $H$ was significantly reduced.

A last thing to note is $\lambda$ of the HLSTM being zero. During training, the HLSTM struggled to get momentum and to start learning, resulting in the hyperparameter search choosing a model with optimal flexibility---a regularisation term of zero. Illustratively, Figure~\ref{fig:losses_all} shows relatively slow HLSTM convergence.

Table~\ref{table:training_settings_complete} shows other training settings (or ``hyperparameters") that were determined through trial-and-error (and not through exhaustive search). All models reduced their learning rates by a factor of 0.1 when the validation loss reached a plateau, had a batch size ($B$) of 16, and used $k$ = 5 in their $k$-fold cross-validation schemes.
$B = 16$ was plainly adopted from \citet{masters2018revisiting}, who found smaller batch sizes $(2 \leq B \leq 32)$ to provide benefits in terms of convergence stability and overall test performance for a given number of epochs.
The MLPs had a patience of 6 and the RNNs of 15 to accomodate for their differences in convergence speed. All activation functions employed are ReLU, including the readout. 

\begin{table}[h]
    \centering
    \caption{List of the hyperparameters included in the grid search.}
    \vspace{-0.75\baselineskip}
    \label{table:hyperparams}
    \begin{tabular}[t]{@{}ll@{}}
        \toprule
        \hspace{0.25cm}\raggedright\emph{Hyperparameter} & \emph{Symbol}\hspace{0.25cm} \\
        \midrule
        \hspace{0.25cm}Hidden layers & $k$ \\
        \hspace{0.25cm}Hidden units & $L^{\kappa}$ \\
        \hspace{0.25cm}Learning rate & $\mu$ \\
        \hspace{0.25cm}Learning rate, shared & $\mu_{\textrm{shared}}$ \\
        \hspace{0.25cm}Learning rate, branches & $\mu_{\textrm{branch}}$ \\
        \hspace{0.25cm}Weight decay & $\lambda$ \\
        \bottomrule
    \end{tabular}
\end{table}

\begin{table}[h] 
    \centering
    \caption{Overview of the used hyperparameters determined through grid search.}
    \label{table:hyperparams_complete}
    \begin{tabular}{lSSSSSS}
        \toprule
        & {$k$} & {$L^{\kappa}$} & {$\mu$} & {$\mu_{\textrm{shared}}$} & {$\mu_{\textrm{branch}}$} & {$\lambda$} \\
        \midrule
        MLP & 4 & 64 & 1e-5 & & & 1e-5 \\
        HMLP & 7 & 64 & & 1e-4 & 7e-4 & 1e-5 \\
        LSTM & 6 & 112 & 1e-3 & & & 1e-6 \\
        HLSTM & 7 & 48 & & 1e-4 & 7e-4 & 0 \\
        GRU & 4 & 128 & 1e-3 & & & 1e-5 \\
        HGRU & 4 & 64 & & 1e-3 & 4e-3 & 1e-7 \\
        \bottomrule
    \end{tabular}
\end{table}

\begin{table}[h] 
    \centering
    \caption{Overview of other used training settings determined through trial-and-error.}
    \label{table:training_settings_complete}
    \begin{tabular}{lllSll}
        \toprule
        & optimizer & $\mu_{\textrm{scheduler}}$ & {patience} & $B$ & $k_{\textrm{folds}}$  \\
        \midrule
        MLP & Adam & ReduceLROnPlateau   & 6 & 16 & 5  \\
        HMLP & Adam & ReduceLROnPlateau  & 6 & 16 & 5  \\
        LSTM & Adam & ReduceLROnPlateau  & 15 & 16 & 5  \\
        HLSTM & Adam & ReduceLROnPlateau & 15 & 16 & 5  \\
        GRU & Adam & ReduceLROnPlateau   & 15 & 16 & 5  \\
        HGRU & Adam & ReduceLROnPlateau  & 15 & 16 & 5  \\
        \bottomrule
    \end{tabular}
\end{table}

\end{document}